\documentclass[11pt]{article}

\usepackage[preprint]{acl}

\usepackage{times}
\usepackage{latexsym}
\usepackage[T1]{fontenc}
\usepackage[utf8]{inputenc}
\usepackage{microtype}
\usepackage{inconsolata}
\usepackage{graphicx}
\usepackage{booktabs}
\usepackage{array}
\usepackage{multirow}
\usepackage[most]{tcolorbox}
\newtcolorbox{promptbox}[2][]{breakable, enhanced, colback=black!4, colframe=black!65,
  boxrule=0.5pt, arc=2pt, left=5pt, right=5pt, top=4pt, bottom=4pt,
  title=#2, fonttitle=\bfseries\small, coltitle=white, colbacktitle=black!65,
  fontupper=\small\ttfamily, #1}
\newcommand{\slot}[1]{\textcolor{blue!55!black}{\{#1\}}}

\title{Policy Central: Detecting Conflicts, Redundancy, and Similarity in Deployed Police Policy}

\author{Holly Baker\thanks{Primary contributor.} \quad Michael Kuo \quad Kyana Pollas \\
  PolicingLab, Miami, Florida, United States \\
  \texttt{\{baker,kuo,pollas\}@policinglab.org}}

\newcommand{\eeAgg}{0.472}
\newcommand{\eeConfl}{0.615}
\newcommand{\eeRedun}{0.561}
\newcommand{\eeSimil}{0.273}

\begin{document}
\maketitle

\begin{abstract}
Policy Central is a deployed system that helps police departments find
relationships between sections of their written policy: whether two
sections \emph{conflict}, are \emph{redundant}, are merely \emph{similar}, or
are \emph{different}. We describe the system and compare two
architectures. \textbf{Retrieve-Judge}, a retrieve-then-classify pipeline,
runs as a cheap library-wide health signal; \textbf{Screen-Confirm}, a screen-then-confirm cascade, is a selective audit of the documents a reviewer
chooses, checking them exhaustively against one another.
On a 302-pair benchmark drawn from one department's
\textasciitilde84 real policy documents, given each pair Screen-Confirm reaches
0.795 aggregate recall and 0.885 on Conflict, so current models judge well.
Finding the pair is the hard part: Retrieve-Judge's composed recall falls to
0.457, and Screen-Confirm recovers \eeAgg{} (\eeConfl{} on Conflict). Today these pairs are found by reading the whole library, nearly
20{,}000 sections, checked against one another and re-checked after every
revision. A scan that returns half of them in minutes, at low cost, and can be
re-run on demand changes the economics of policy review even before recall
improves.
\end{abstract}

\section{Introduction}

Police departments run on written policy: standard operating procedures,
general orders, and directives, written by each agency for itself \citep{useofforce-policies} and edited over many years. The result routinely
contains internal tension: one order requires what another forbids, or two
procedures cover the same ground in different words. Finding them by hand is slow, and getting it wrong has real consequences in the
field.

We introduce Policy Central, a system in pilot at one U.S.\ police department with
broader rollout underway, which surfaces these relationships automatically. It classifies pairs of policy sections into one of four relationships:
\textbf{Conflict} (the sections cannot both be followed), \textbf{Redundancy}
(they say the same thing and one is unnecessary), \textbf{Similarity} (they are
related but not redundant), or \textbf{Different} (unrelated). Conflict is the
high-value, low-frequency case; most pairs in a real library are Different.

Policy Central serves two operational needs with two architectures. Given the pair,
current models judge well, Conflict included; when they must find the pair themselves, both architectures
lose much of that recall, in different places. This paper reports where, and what it
took to recover some of it. Our contributions:

\begin{enumerate}
  \item A deployed system with two architectures, \textbf{Retrieve-Judge} and
  \textbf{Screen-Confirm}: the first a retrieve-then-classify pipeline run as a
  library-wide health signal, the second a screen-then-confirm cascade run as a selective audit of the documents a reviewer chooses.
  \item A 302-pair benchmark of real policy pairs, on which we compare both
  architectures and two baselines (Llama zero-shot, DeBERTa NLI), given the pair and
  having to find it, and report the production cost and time of both modes.
\end{enumerate}

\section{Related Work}

\begin{figure*}[t]
  \centering
  \includegraphics[width=0.90\textwidth]{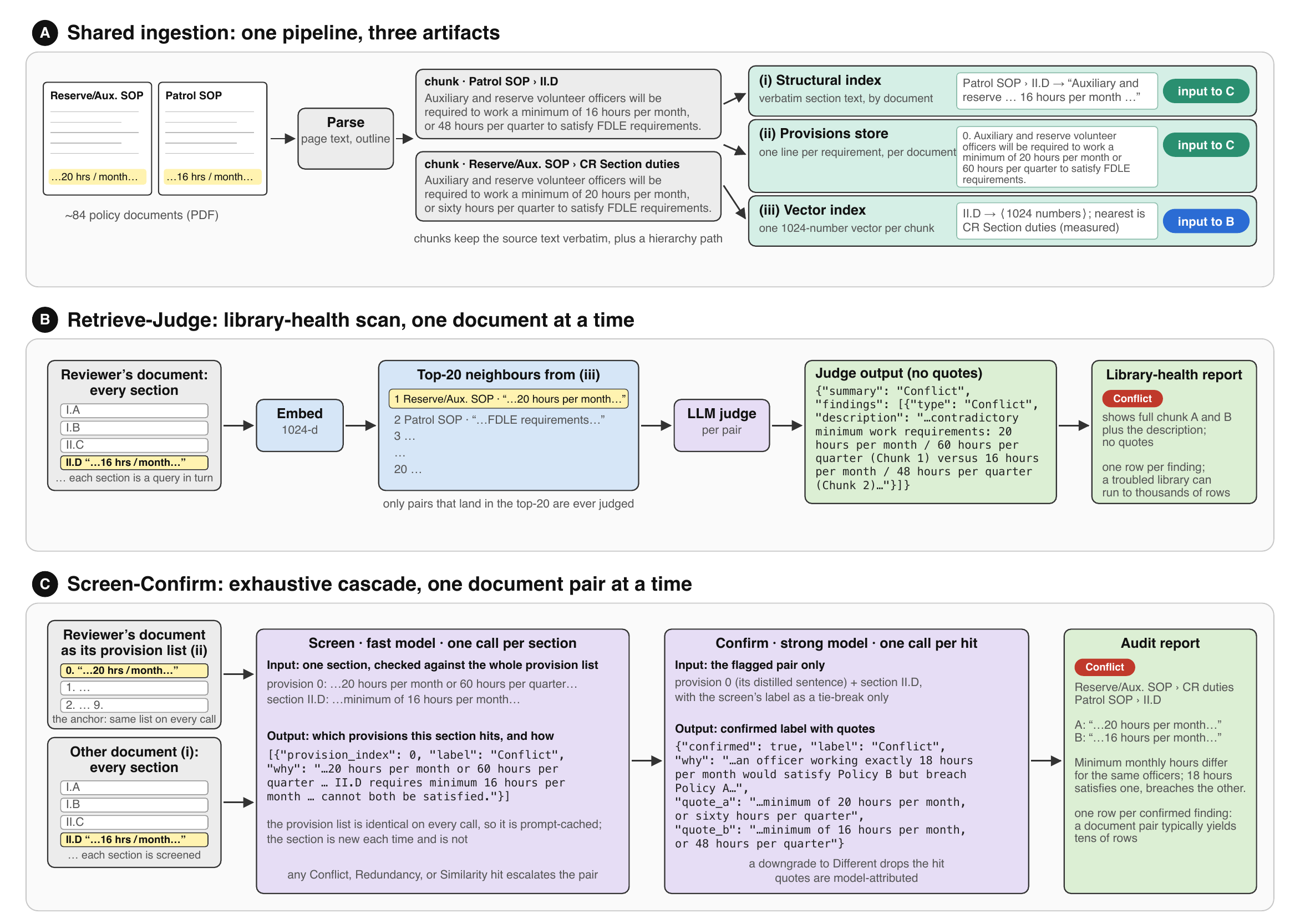}
  \caption{Policy Central, traced on one real benchmark pair (hp-0026,
  Appendix~\ref{sec:appendix-benchmark}). One ingestion pipeline (A) writes
  three artifacts. \textbf{Retrieve-Judge} (B) reads only the vector index and
  judges each section's top-$k$ neighbours; for this pair, the partner section is the nearest neighbour.
  \textbf{Screen-Confirm} (C) reads the structural index and provisions store,
  never the vector index, and sweeps every section through a cheap-screen /
  expensive-confirm cascade. Every model output shown is the scored pipeline's
  actual output, abbreviated with ``\ldots''.}
  \label{fig:architecture}
\end{figure*}

\textbf{The problem.} Legal and regulatory text grows steadily \citep{coupette}, and with it the room for contradiction.
Policing shows it at scale: a
review of over 11,000 pages of use-of-force policy from the 100 largest U.S.\
cities found policies varying widely in structure and length, some departments spreading their force rules across multiple sections of their general orders
\citep{useofforce-policies}, and a systematic review of NLP in policing covers
no work on policy documents themselves \citep{policing-nlp}.

\textbf{Prior systems.} Approaches range from formal machinery to text-level methods.
Knowledge-graph systems detect conflict over a structured representation,
extracted automatically from text in PoliGraph \citep{poligraph} or hand-encoded
as deontic axioms \citep{deontic-kg}. Benchmark work targets
conflict directly: LegalWiz generates legal-style documents with injected
contradictions \citep{legalwiz}, and Clash-of-Leges and a Chinese legal-conflict
dataset benchmark statutory conflict detection \citep{clashofleges,lcrcn}. Applied compliance systems work against an external standard rather than
within one library: a procurement-audit framework checks purchase records
against compliance policy on real procurement data \citep{cocaf}, and a
maintenance-manual retrieval system keeps every answer traceable to certified
sources \citep{aircraft-mro-compliance}. Neither judges whether two sections of
one library conflict. Duplication in the U.S.\ Code \citep{coupette-dedup} and
an open-source NLI pipeline over U.S.\ defence policy documents
\citep{policy-contradictions-nlp} measure relations directly.

\textbf{Framing.} Natural language inference is the natural framing \citep{contractnli,docnli}, adapted in Section~\ref{sec:taxonomy}. Conflict is hard for models: even GPT-4 is unreliable on self-contradictions
that need context to see \citep{contradoc}, and current LLMs fail to detect
conflicts across clinical guidelines \citep{neurosymbolic-multimorbidity}. Our two
architectures rest on existing patterns: Screen-Confirm is a cost cascade
\citep{frugalgpt,gatekeeper,babybear,violajones}; Retrieve-Judge combines similarity retrieval \citep{bm25,mteb} with a
prompt-based LLM judge, as in \citet{schumann-regulatory}. Retrieving by
similarity alone can miss what a task needs \citep{metrag}, and a rule that
conflicts with another need not read like it.

\section{System Design}
\label{sec:system}

A reviewer picks a policy document and the system reports which sections
elsewhere in the library relate to it, each pair labelled with a one-line
reason. Two architectures produce that report. \emph{Retrieve-Judge} embeds
every section, retrieves its nearest neighbours, and judges each pair
(panel~B of Figure~\ref{fig:architecture}, ending in the library-health
report). \emph{Screen-Confirm} checks every section of another document (the
\emph{compared} document) against every requirement of the reviewer's (the
\emph{anchor}), cheaply first and carefully on what survives (panel~C, ending in the audit report). Both read the artifacts
that one ingestion pipeline writes (panel~A).

\subsection{Relation Taxonomy}
\label{sec:taxonomy}
Table~\ref{tab:taxonomy} defines the four labels with real excerpt pairs; each
maps to a different reviewer action.

\begin{table}[t]
  \centering
  \small
  \setlength{\tabcolsep}{4pt}
  \renewcommand{\arraystretch}{1.05}
  \begin{tabular}{@{}>{\raggedright\arraybackslash}p{1.75cm}>{\raggedright\arraybackslash}p{2.2cm}>{\raggedright\arraybackslash}p{3.15cm}@{}}
    \toprule
    \textbf{Label} & \textbf{Definition} & \textbf{Example (from the corpus)} \\
    \midrule
    \textbf{Conflict}
      & Requirements that cannot both be satisfied.
      & \emph{``Must attend a two hour refresher training annually.''} vs.\
        \emph{``Must attend a four (4) hour refresher training annually.''} \\
    \addlinespace[2pt]
    \textbf{Redundancy}
      & Same requirement; one section is deletable with no loss.
      & \emph{``In sensitive criminal matters, the Unit Commander shall be
        notified prior to the reopening of an investigation''}, in two units'
        SOPs. \\
    \addlinespace[2pt]
    \textbf{Similarity}
      & Related; a reader needs both, but neither subsumes the other.
      & \emph{``\ldots kept secret unless \ldots duty requires
        otherwise''} vs.\ \emph{``At no time \ldots release any information
        considered to be confidential.''} \\
    \addlinespace[2pt]
    \textbf{Different}
      & No operationally relevant relation.
      & \emph{``Being armed in any manner.''} vs.\ \emph{``Must be a minimum
        of 14 years of age.''} \\
    \bottomrule
  \end{tabular}
  \caption{The four relation labels, each with a verbatim excerpt pair from
  the pilot department's corpus that the deployed system labelled correctly
  (provenance in Appendix~\ref{sec:appendix-benchmark}).}
  \label{tab:taxonomy}
\end{table}

The product exists to surface \textbf{Conflict}, because two sections that
cannot both be followed mean the library is wrong somewhere. \textbf{Redundancy}
matters for upkeep, since a requirement stated twice can be cut to one copy so
that future edits happen once. \textbf{Similarity} exists so that a reviewer
drafting or revising one policy sees the related sections elsewhere in the
library, to reference or reconcile rather than delete. Redundancy and Similarity are separated by a deletion test: if one section can
go because the other carries its content, the pair is Redundant; if both must
stay, it is Similar. Similarity versus Different is likewise a question about the reviewer, not the
text: two sections are Similar if a reviewer revising one would need to read the
other, and Different if not, however much wording they share.

The four labels are natural language inference (NLI) labels adapted to what a
reviewer does next. Building on NLI gives
``contradicts'' a precise meaning, connects to prior work on legal documents
\citep{contractnli,docnli}, and supplies trained models as a free baseline. The
mapping is: contradiction $\rightarrow$ Conflict; entailment in both directions
$\rightarrow$ Redundancy; neutral $\rightarrow$ Similarity \emph{or}
Different, split by whether a reviewer must act. Conflict drops NLI's
direction: NLI asks whether a premise contradicts a hypothesis, whereas two
rules that cannot both be followed conflict whichever is read first. The last
line is where policy review departs from NLI, since two of our labels share one
NLI label. An off-the-shelf NLI model can therefore
find Conflict at best and collapses the rest, which is what DeBERTa NLI shows
(Section~\ref{sec:eval-results}): 0.654 Conflict recall and near zero on
Redundancy and Similarity.

\subsection{Data Ingestion}
\label{sec:ingest}
A single pipeline (panel~A) serves both architectures but writes \emph{three}
artifacts, each read by a different consumer. The parse (page-by-page extraction, Unicode normalisation, and
outline-aware segmentation that records each chunk's \texttt{hierarchy\_path}) is shown in the figure.
Chunking is a deterministic length-based split (${\sim}$600 tokens) that keeps
the source text verbatim.

The artifacts are: (i)~a
\textbf{structural index}: raw chunk text and \texttt{hierarchy\_path},
no embeddings, read by Screen-Confirm for its \emph{compared} document;
(ii)~a \textbf{distilled provisions store}, read by Screen-Confirm for its
\emph{anchor} document; and (iii)~a \textbf{vector index} (Qdrant): each leaf chunk embedded once (Section~\ref{sec:rj}) for
similarity search, read \emph{only} by Retrieve-Judge. The provisions store is
built by a separate ingest-time LLM pass (Claude Haiku, once per document) that
rewrites a document's chunks into one standalone sentence per requirement
(prompt in Appendix~\ref{sec:appendix-prompt-distill}), each grounded back to its
source chunk, an obligation-level analogue of proposition retrieval units
\citep{dense-x} and atomic facts \citep{factscore}. It is cached and reused by
every later scan. A requirement it fails to extract is invisible on the anchor
side; Section~\ref{sec:eval-e2e} measures that loss.

Retrieve-Judge's retrieval unit is the leaf chunk (19{,}737 across the
department's ${\sim}84$ documents, Table~\ref{tab:settings}); Screen-Confirm
judges whole sections, so the two compare different-sized units (Limitations).

\subsection{Retrieve-Judge}
\label{sec:rj}
Retrieve-Judge (panel~B) takes each leaf chunk of the reviewer's document in turn: each is embedded, its top-$k$
nearest neighbours by embedding similarity, drawn from every document in the
library, are retrieved, and each (chunk, neighbour) pair is judged. Chunks are
embedded with Cohere Embed v4 via Bedrock (1024 dimensions). The deployed audit retrieves
$k{=}20$ under cosine distance and keeps neighbours above a 0.75 similarity
threshold. The judge returns one JSON object per pair, \texttt{\{summary,
findings[\{type, description\}]\}}: a label and a one-sentence description that
must name the relation (``deadline is 4 hours in A versus 24 hours in B''), with
no reasoning trace and no quotations (Appendix~\ref{sec:appendix-prompt-text}).
Retrieval is dense-only; a reranker, hypothetical-document queries, BM25,
embedding-side changes, and a larger $k$ were all tried and none moved recall
(Appendix~\ref{sec:appendix-negative}), and Section~\ref{sec:eval-results}
shows that judge swaps do not either. The constraint is structural: Conflict is
a \emph{logical} relation, not a similarity one.
Similarity surfaces contradictions that share wording: the 20-hour and 16-hour
minimums in Figure~\ref{fig:architecture} are each other's top-1 neighbour. It
has no signal for conflicts that share no wording: a rule that narcotics
officers never wear uniform off duty and one that uniform must be worn on
extra-duty jobs conflict, yet neither is in the other's top-20 (a benchmark
pair). Retrieve-Judge therefore samples the library rather than covering it, and we
use it as a library-health scan rather than an audit (Limitations).

\subsection{Screen-Confirm}
Screen-Confirm (panel~C) replaces similarity sampling with an
exhaustive two-stage cascade: discard the Different majority cheaply, reason
carefully about the survivors, the Viola--Jones intuition \citep{violajones}
as a cost cascade \citep{cascade-ranking}.
No model ever sees a whole PDF. The compared document is walked one section at a
time, and each section is matched against the anchor document's \emph{provision
list}, its distilled one-line-per-requirement form, rather than its raw text.
Each screen call is therefore short, and the provisions block is the same on
every call, so it can be prompt-cached. Headings, definitions, and repeated
administrative language that distillation left out are never sent to the model.

\textbf{Screen.} One Claude Haiku call per compared-document section receives the numbered
provisions list plus that section and returns a JSON array of candidate hits (as in the
figure). There is no score or threshold: any hit labelled Conflict, Redundancy,
or Similarity escalates that provision--section pair, and only Different does not.

\textbf{Confirm.} Each escalated pair goes to one Claude Sonnet call that sees
three things: the flagged provision as its distilled sentence rather than its
source chunk, the compared document's raw section, and the screen's label,
which it may use only as a tie-breaker. It returns a confirmed label with \texttt{quote\_a}
and \texttt{quote\_b}; a downgrade to Different drops the hit.
\texttt{quote\_a} is therefore drawn from the distilled sentence and
\texttt{quote\_b} from the section, and neither is code-validated against the
source, so we call them model-attributed. Confirm does not cache (each pair is unique); findings render as an
audit report (panel~C), typically tens of rows for a document pair.

A scan costs $N$ screen calls plus $M$ confirm calls ($N$ sections, $M$
escalations); distillation is a once-per-document ingest cost. 

\section{Evaluation}
\label{sec:eval}

\subsection{Benchmark and Its Construction}
\label{sec:eval-benchmark}
Our benchmark is 302 pairs (26 Conflict, 57 Redundancy, 44 Similarity, 175
Different), a distribution that reflects real policy libraries, where
Different is the large majority. Every
pair is two real sections from the pilot department's policies. Labels are \emph{semi-automatic} rather than subject-matter expert: a Haiku
pass extracts checkable claims per document and a pure-code stage (no LLM)
pairs them, a \emph{non-Claude} judge (GPT-5.6 Sol) labels each pair, an Opus
model attacks each label, and people check the disputed rows, following the
extract-then-verify pattern of \citet{mizanqa} (Figure~\ref{fig:benchmark},
Appendix~\ref{sec:appendix-benchmark}).

We do not field the mining pipeline as a detector, since scoring it on a
benchmark it produced would be circular; one dependency remains, in that its
claim extraction and production distillation are both Haiku passes
(Limitations).

\subsection{Systems and Methodology}
\label{sec:eval-methodology}
We compare four systems on the 302 pairs. \textbf{Screen-Confirm}
and \textbf{Retrieve-Judge} are as described in Section~\ref{sec:system}.
\textbf{Llama zero-shot} is a single call with a four-line prompt, no examples and
no retrieval, a floor for judgment. \textbf{DeBERTa NLI} is an off-the-shelf
zero-shot NLI checkpoint, not fine-tuned on our data
(Appendix~\ref{sec:appendix-prompt-baselines}).

Table~\ref{tab:perclass-recall} asks two questions. The lower block asks how well
each system judges a pair it is handed; the upper block asks how much of that
survives when the system must find the pair itself. Each architecture appears in
both, so the gap between its two rows is what finding costs it. Table~\ref{tab:settings} lists what each system sees in each setting. Handed
the pair, the cascade's screen still has to pick the right provision, but out of
16 rather than out of the whole anchor, and Retrieve-Judge's judge simply reads
the two sections. To find the pair, Retrieve-Judge must first retrieve it, so we estimate its
recall ($R$, the fraction of gold pairs caught) as
$R_{\text{find}} = R_{\text{retrieve}} \times R_{\text{judge}}$, where
$R_{\text{retrieve}}$ is the fraction of gold pairs whose partner is in the top
10 and $R_{\text{judge}}$ the fraction the judge labels correctly when handed
the pair; for Conflict, $0.571 \times 0.731 = 0.418$. Multiplying assumes the
two are independent (Limitations). Retrieval recall is measured on the 137
pairs whose two quotes both map to an indexed leaf chunk (14 Conflict, 13
Redundancy, 25 Similarity), each queried in both directions; judge recall is
over every pair of the class. Screen-Confirm's end-to-end recall is measured
directly, one pass per pair, but reads the same way:
$R_{\text{e2e}} = R_{\text{reach}} \times R_{\text{screen}} \times R_{\text{confirm}}$,
the fractions of gold positives that reach the screen, that the screen then
escalates, and that confirm then labels correctly. Over all classes that is
$0.945 \times 0.742 \times 0.674 = \eeAgg$
(Appendix~\ref{sec:eval-evolution} gives the counts).

\begin{table}[t]
  \centering
  \footnotesize
  \setlength{\tabcolsep}{3pt}
  \renewcommand{\arraystretch}{1.1}
  \begin{tabular}{@{}>{\raggedright\arraybackslash}p{1.35cm}>{\raggedright\arraybackslash}p{3.0cm}>{\raggedright\arraybackslash}p{2.75cm}@{}}
    \toprule
    & \textbf{Screen-Confirm} & \textbf{Retrieve-Judge} \\
    \midrule
    \textbf{Given the pair}
      & Screen: the gold provision and 15 others from its document, against the gold section. Confirm: each hit.
      & Judge: the two gold sections. \\
    \addlinespace[2pt]
    \textbf{Must find the pair}
      & Screen: every anchor provision (mean 614, max 13{,}352) against every section of the compared document. Confirm: each hit.
      & Retrieve: nearest of 19{,}737 chunks by cosine ($k{=}20$ in deployment; recall scored at 10). Judge: each retrieved pair. \\
    \bottomrule
  \end{tabular}
  \caption{What each system is shown in the two settings of
  Table~\ref{tab:perclass-recall}.}
  \label{tab:settings}
\end{table} All model calls use temperature 0 and pinned
model IDs; prompts are in Appendix~\ref{sec:appendix-prompt-text}.

\begin{table}[t]
  \centering
  \scriptsize
  \setlength{\tabcolsep}{2.5pt}
  \begin{tabular}{@{}lcccc@{}}
    \toprule
    \textbf{System} & \textbf{Confl.} & \textbf{Redun.} & \textbf{Simil.} & \textbf{Agg.$^\dagger$} \\
    & (26) & (57) & (44) & (127) \\
    \midrule
    \multicolumn{5}{@{}l}{\emph{Must find the pair}} \\
    Retrieve-Judge, retrieval only & 0.571 & 0.846 & 0.480 & 0.663 \\
    Retrieve-Judge, Sonnet 4.5 judge & 0.418 & 0.668 & 0.207 & 0.457 \\
    Retrieve-Judge, Llama judge & 0.462 & 0.757 & 0.262 & 0.525 \\
    Screen-Confirm, end-to-end$^\ast$ & \eeConfl & \eeRedun & \eeSimil & \eeAgg \\
    \midrule
    \multicolumn{5}{@{}l}{\emph{Given the pair}} \\
    Retrieve-Judge judge, Sonnet 4.5 & 0.731 & 0.789 & 0.432 & 0.654 \\
    Screen-Confirm                 & \textbf{0.885} & 0.842 & 0.682 & \textbf{0.795} \\
    Llama 3.3 70B zero-shot        & \textbf{0.885} & \textbf{0.947} & 0.455 & 0.764 \\
    DeBERTa NLI, off the shelf & 0.654 & 0.000 & 0.023 & 0.142 \\
    \bottomrule
  \end{tabular}
  \caption{Recall by relation class. $^\dagger$Support-weighted recall over
  Conflict, Redundancy, Similarity ($n=127$); Different is excluded, since
  predicting Different for everything would score well on it while catching
  nothing. Upper block: the system must find the partner section. Retrieve-Judge
  rows are retrieval recall (R@10, over the 137 pairs whose quotes map to indexed
  leaf chunks, Section~\ref{sec:eval-methodology}) alone, then retrieval times
  judge recall for two of eight judges (Appendix~\ref{sec:appendix-leaderboard});
  $^\ast$measured on the
  production data path. Lower block: the system is handed the pair. Confl., Redun.,
  Simil.: Conflict, Redundancy, Similarity.}
  \label{tab:perclass-recall}
\end{table}

\subsection{Results by Relation Class}
\label{sec:eval-results}
\label{sec:eval-e2e}
Given the pair, judgment is largely solved: a plain zero-shot Llama reaches 0.764
and Screen-Confirm 0.795, with the cascade's gain concentrated on Similarity
(0.682 vs.\ 0.455), and Retrieve-Judge's own judge 0.654. Having to find the
pair is what costs, and the end-to-end figures should be read against the
alternative, a reviewer reading the library. We did not choose this recall. The cascade is tuned to over-flag rather than
miss, so its low precision is by design, but every missed pair is a loss, and
Appendix~\ref{sec:eval-evolution} shows at which stage each one was lost. The rest of this section reports what we tried to raise it and what did not move.
Retrieve-Judge falls from 0.654 to 0.457, and the loss is
retrieval: its retrieval alone recovers 0.663, swapping its judge across eight
models moves the composed figure by only about 0.12, and replacing the dense
retriever with BM25 barely moves it (0.455 against 0.468 on the same pairs). Screen-Confirm falls
from 0.795 to \eeAgg{} and pays at its screen (Appendix~\ref{sec:eval-evolution}); on
Conflict it keeps \eeConfl{}, above Retrieve-Judge's 0.418--0.462.

Read by class, the pattern is consistent. Redundancy is the easy class
everywhere: duplicated text shares wording, so retrieval reaches 0.846 and most
judges score it highest; when the cascade loses a Redundancy pair end to end,
the largest single cause is confirm calling it Similarity (8 of 25 misses). Conflict is harder to find than to judge: embeddings rank by how alike two
chunks are overall, and two rules that contradict each other differ at one
detail while the chunk holding the partner rule often sits among unrelated
requirements, so a cosine neighbourhood ranks it behind chunks that merely
discuss the same topic. Retrieval reaches only 0.571, and end-to-end the
cascade keeps \eeConfl{}, with the screen the larger loss; handed the pair,
the capable judges reach 0.73 to 0.89. Similarity is the hard class in every setting (retrieval 0.480,
judges 0.43 to 0.68, end-to-end 0.273), because it is the residual class, bounded on one side by Redundancy and on the other by
Different, so every call on it is a call on two judgment boundaries at once,
and confirm's errors go both ways (5 Similarity pairs read as Redundancy, 8 as
Conflict). 

Swapping the confirm model did not help: every alternative lowered Conflict
recall or lost the ability to walk a false escalation back to Different
(Table~\ref{tab:confirm-swap}, Appendix~\ref{sec:appendix-leaderboard}).

\subsection{Time and Cost on the Department's Library}
\label{sec:eval-cost}
Table~\ref{tab:scale-cost} gives what the two modes cost on the 78 of the department's 84 documents that
were indexed when these runs were made (19{,}646 sections, Screen-Confirm's unit), from production run records and
Bedrock's token metrics. Retrieve-Judge has audited the whole library: one pass
judges the roughly 4{,}800 neighbour pairs above the 0.75 threshold for \$13 in
five minutes, with the judge's static prompt cached (8.9M of 10.4M input tokens
were cache reads). Screen-Confirm has run only as targeted scans of one document
against one or two others, at list price because a provision store that small
is below the screen model's cacheable minimum; scaled by sections, one SOP
against the whole library would take about an hour at ten workers and cost
\$40 to \$80. Full-library scans are not the intended use: in the pilot,
reviewers told us their main use is comparing a document against the few
documents they are revising or need to align, which is the targeted-scan case in
Table~\ref{tab:scale-cost}. Cheaper models would cut cost further: zero-shot Llama 3.3 70B scores 0.764 given
the pair against Screen-Confirm's 0.795 at about one two-hundredth of the cost
(Appendix~\ref{sec:appendix-cost}). The two modes therefore
serve different cases: the thresholded audit is the cheap library-wide signal,
the exhaustive scan the per-pair check, and at today's recall either gives a
reviewer a starting point that reading the library does not.

\begin{table}[tbp]
  \centering
  \scriptsize
  \setlength{\tabcolsep}{2.5pt}
  \begin{tabular}{@{}lrrr@{}}
    \toprule
    \textbf{Scan} & \textbf{Calls} & \textbf{Cost} & \textbf{Time} \\
    \midrule
    \multicolumn{4}{@{}l}{\emph{Screen-Confirm, measured targeted scans}} \\
    SOP (54 prov.) vs.\ 2 SOPs, 885 sec. & 885+216 & \$1.60 & 150\,s \\
    SOP (56 prov.) vs.\ 1 SOP, 100 sec. & 100+66 & \$0.41 & 42\,s \\
    \multicolumn{4}{@{}l}{\emph{Screen-Confirm, projected}} \\
    One SOP vs.\ library, 19.6k sec. & 25--33k & \$40--80 & ${\sim}$1\,h \\
    \midrule
    \multicolumn{4}{@{}l}{\emph{Retrieve-Judge, measured audit}} \\
    Library audit, 4{,}807 pairs judged & 5.2k & \$12.86 & 4.8\,min \\
    \bottomrule
  \end{tabular}
  \caption{Cost and wall-clock of production scans on the deployment (ten
  scan workers). Calls are screen\,+\,confirm for Screen-Confirm and judge
  calls for Retrieve-Judge; times are from the run records; costs are Bedrock's
  logged token counts for each run at list rate, cache reads at the cache rate.
  The projected row scales the measured per-section cost and time; it is an upper
  bound, not the intended use.}
  \label{tab:scale-cost}
\end{table}

\section{Conclusion and Future Work}
Finding rules that contradict or repeat one another is currently a manual review
task. Policy Central does not replace that review, but it gives the reviewer a
report of candidate pairs to start from.
Conflict is a logical relation, not a similarity, so a retrieve-then-judge pipeline has a
ceiling that none of the judge or retriever swaps we tried lifts. Handed a pair, current models label it
well, and the losses come from reaching the pair. An exhaustive cascade over
distilled requirements fits the operational need: it trades precision for recall, since a
reviewer dismisses a false flag quickly but never sees a missed one; its cost is linear
in the number of sections, low enough to re-run after each edit; and it reports on a
document pair in minutes, so a reviewer can fix and re-check in one sitting.

In deployment, the next steps are improving finding
quality against reviewer feedback from the pilot, extending the rollout under way to
further departments, and reporting usage data. Ingestion is the main barrier to other corpora:
the parser is pattern-based and tuned to one department's directive formats, and
layout-aware document models \citep{layoutlm,donut} are one possible route to
ingesting any agency's PDFs unchanged. Beyond policing, other rule-bound
domains report the same inconsistency problem: clinical guidelines \citep{medical-guidelines}, contracts
\citep{contract-ambiguity}, and immigration rules
\citep{immigration-inconsistency}. Recall matters most for a possible step
after detection: an agent that runs the scan, resolves the clear cases, and
brings the reviewer the edge cases, an extension that Ethical Considerations
flags for separate review. Rewrites
reopen evaluation, since a proposed text must be graded for grammaticality, naturalness,
and completeness as in the QA2D evaluation of \citet{qa2d}.

\section*{Limitations}

\textbf{Benchmark.} The most important limitation is label provenance: the
302 pairs are real, but their labels are semi-automatic, not subject-matter
expert. Our SME annotators did label an earlier pool, whose underlying pairs
turned out too narrowly distributed to report; the annotators and protocol are
in place and labelling this benchmark is the next step. We have no
inter-annotator agreement figure for this benchmark. The label set has no
``partly redundant'' class, so pairs that share only part of their content are
forced to Redundancy or Similarity; a fifth label would only move that boundary. The benchmark is also one
department's library: 302 pairs from a single corpus of about 84 documents, with
26 Conflicts, so the Conflict figures carry wide intervals
(Appendix~\ref{sec:appendix-precision}) and transfer to other departments'
formats and drafting conventions is untested. The end-to-end Conflict recall of
\eeConfl{} is a reported finding about the production path, not a property of
the models; Appendix~\ref{sec:eval-evolution} attributes the shortfall stage by stage.

\textbf{Development and evaluation share one benchmark.} The definition fix, the
prompt iterations and two design fixes of Appendix~\ref{sec:eval-evolution} were found and measured on the same 302 pairs, with no
held-out split. Absolute recall is therefore optimistic, and the number to trust is the
direction of each change rather than its size. The model-swap and retriever ablations
vary one component with everything else fixed, so they are relative comparisons and less
exposed. Every model-swap result was obtained under prompts tuned against Sonnet 4.5, so
swaps measure prompt transfer as much as model capability. The SME labelling above will
provide the held-out set.

\textbf{Measurement.} Retrieve-Judge judges leaf chunks and Screen-Confirm whole
sections, so unit size is an uncontrolled variable in the comparison. Parsing
errors cascade \citep{doc-parsing-unveiled}, and here they are silent: a wrong
section boundary changes what confirm reads and can turn a correct judgment into
a wrong one with no trace in the output. Our stage-by-stage attribution of misses (Appendix~\ref{sec:eval-evolution})
checks that gold text is present in the index and that a section can be
located, not that boundaries are right,
and we report no boundary-level accuracy for the parser; the three ingest faults
were found only because that attribution placed misses before any model call.
The cascade itself is a design choice with a counter-argument: escalation pays
only when the screen's decision to escalate carries information about whether
confirm will be right. Our screen escalates 36 to 71\% of sections in the
end-to-end run depending on anchor size, and the gain we measure over a single
pass is in recall, not in cost per correct finding, which we have not measured. Distillation
recall is a lower bound: we ask whether each gold quote is contained in some provision,
a token-overlap test, so a requirement the distiller restated in its own words counts as
missed. The misses we did see are mostly descriptive statements that a
requirement-extraction prompt does not treat as rules, such as \emph{``Sworn personnel
work ten hours a day, four days a week''}, a Redundancy target in the benchmark that never
became a provision. Confirm also judges the distilled sentence rather than the
source chunk it was distilled from, whereas the given-pair test handed it source
text; each provision retains its chunk id, so passing the chunk to confirm is a
one-line change that would also make \texttt{quote\_a} verbatim from the
document, and we have not yet measured it.

\textbf{What the benchmark cannot see.} Benchmark candidates were mined by a Claude
Haiku pass, and production distillation is also a Haiku pass, with separate code and
prompts. A kind of requirement that Haiku reliably fails to extract, for instance one
stated as a schedule or in a table rather than as a ``shall'', would be missing from the
benchmark and from the provisions store alike, so the benchmark would score the system as
complete where it is not. Candidates were also paired by shared terms, so the
benchmark contains no pair with zero lexical overlap, and Retrieve-Judge's
retrieval gap is measured on pairs that do share wording. And because the benchmark is enriched for the rare classes,
precision will be lower in deployment, while recall, the figure we report, does not
depend on how rare a class is (Appendix~\ref{sec:appendix-precision} works the
numbers).

\textbf{Retrieval churn.} A distinct limitation concerns using Retrieve-Judge \emph{repeatedly} as a
library-health check. Because its top-$k$ retrieval is corpus-state-dependent,
re-running an audit after document edits changes the embeddings and therefore the
neighbor sets: fixed pairs can drop out before their resolution is confirmed,
genuinely new pairs surface, and the reported issue count can stay flat or even rise
despite real improvement, purely from retrieval churn rather than any statement about
the underlying corpus. This is a known property to design around, and it is the main
reason we route iterative, check-your-work usage to Screen-Confirm, whose exhaustive
per-section screen re-checks the full space on every run and so drops reliably as
issues are fixed.

\textbf{Composed estimates.} Retrieve-Judge's find-the-pair figures are a
product of two separately measured quantities, retrieval recall and judge recall
given retrieval, which assumes the two difficulties are independent for a given
pair, a simplification prior RAG work has named \citep{stochastic-rag}. If they
are correlated, for instance if ambiguous pairs are both harder to retrieve and
harder to judge, the true figure could sit on either side of the estimate, and
retrieval metrics are known to track downstream performance only weakly
\citep{erag}. The retrieval factor is measured only on the 137 pairs whose
quotes could be mapped to indexed leaf chunks, so the composition applies a
subset's retrieval rate to the whole class's judge rate. The estimate also uses
recall at 10 with no score threshold, whereas the production audit keeps only
neighbours above 0.75 cosine, so it is an upper bound on what the deployed audit
retrieves. Measuring Retrieve-Judge
end-to-end in one pass, as we do for Screen-Confirm, is the next measurement
step.

\section*{Ethical Considerations}
Policy Central operates on police policy, a sensitive domain. The system is
advisory: it surfaces candidate relationships for human policy reviewers and
never edits, approves, or enforces policy. Whether two policies in fact
conflict, and what to do about it, remains a determination for the agency's
policy and legal personnel; the system only proposes. Every finding carries
model-attributed quotes so a reviewer can check it against the source text,
and reviewers are told that Conflict findings are low-precision (0.404) by
design. The failure mode that matters more is the false negative: on the
production path the system recovers \eeConfl{} of benchmark Conflicts, so a
clean report is not evidence that a library is clean, and the tool is presented
to departments as a complement to review, not a certification of it. The policies are public records of the department, and the department
reviewed Policy Central's output and gave feedback throughout the pilot. Policy
text is processed within the operating organisation's own AWS environment. The only policy text reproduced here is the short excerpts in
Table~\ref{tab:taxonomy}, Figure~\ref{fig:architecture}, and
Appendix~\ref{sec:appendix-benchmark}, with document identifiers removed. The benchmark was labeled semi-automatically rather than by
subject-matter experts, so its labels are not an authoritative reading of any
policy. Finally, a tool that flags policy conflicts could be misread as
evidence about an individual officer's compliance; the system is scoped to
document review and is neither designed nor validated for adjudicating
individual conduct. Systems built for a narrow advisory purpose tend to expand
into adjacent decision areas without a corresponding reassessment, so any
extension of Policy Central toward proposing or applying fixes, or toward
enforcement-relevant determinations, should trigger a fresh risk assessment
rather than inherit this one.

\bibliography{references}

\appendix
\section{Benchmark Construction and Example Provenance}
\label{sec:appendix-benchmark}

\begin{figure*}[t]
  \centering
  \includegraphics[width=\textwidth]{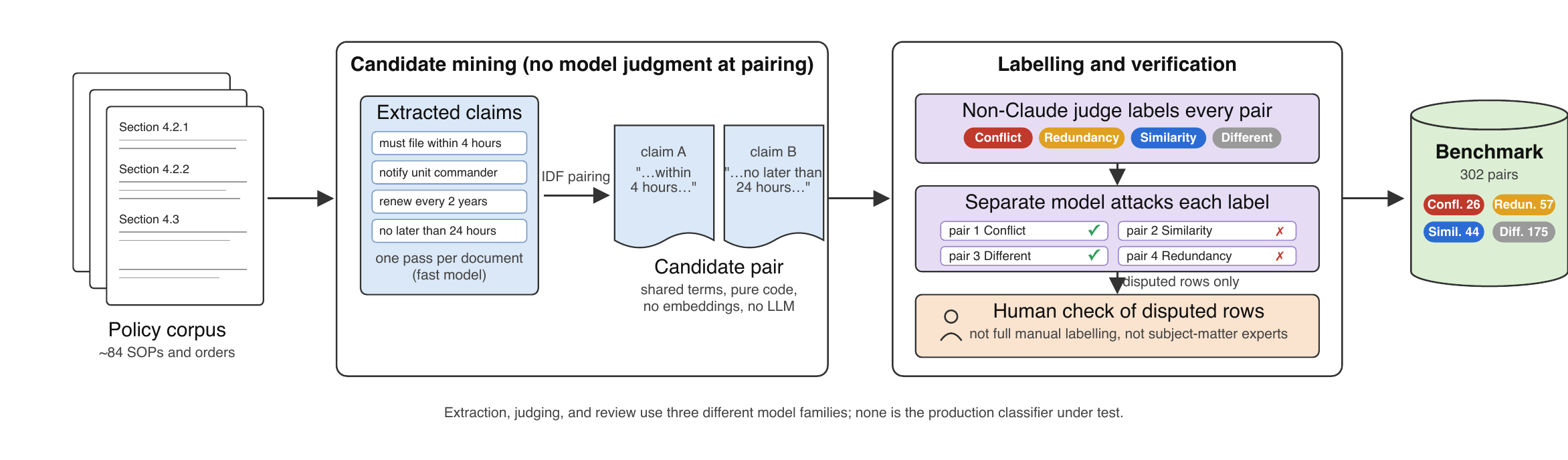}
  \caption{Benchmark construction. Claims are extracted per document and paired
  by shared terms with no model judgment; every pair is labelled by a non-Claude
  judge, each label is attacked by a separate model, and only disputed rows are
  hand-checked; the hand-checked label is final and replaces the judge's.}
  \label{fig:benchmark}
\end{figure*}

Figure~\ref{fig:benchmark} shows the four-stage construction of the 302-pair
benchmark described in Section~\ref{sec:eval-benchmark}: mechanical candidate
mining with no model judgment at the pairing step, labelling by a non-Claude
judge, an adversarial review pass by a separate model, and human verification
of the disputed subset only. Candidates entered the pool three ways: a stratified
random sample of extracted claim pairs (174), a value-mismatch gate that pairs
claims on one topic carrying different numbers or thresholds (75), and a
same-topic gate that pairs claims restating one value (53), which supplies most
Redundancy candidates. Every passage is verbatim document text; only the
pairing step differs. 33 of the 57 Redundancy pairs are character-identical
text appearing in two places.

\section{Judge-Model Leaderboard}
\label{sec:appendix-leaderboard}

Table~\ref{tab:leaderboard} gives the full Retrieve-Judge leaderboard across the
eight judge models we evaluated. The headline point is the tight clustering:
every model lands between 0.46 and 0.57 aggregate recall, a spread of about
0.12, so the choice of judge model moves Retrieve-Judge far less than the
architecture does. This is the
evidence behind the claim in Section~\ref{sec:eval} that the ceiling is intrinsic
to the retrieve-then-judge architecture rather than to any one model.

\begin{table}[ht]
  \centering
  \small
  \begin{tabular}{lc}
    \toprule
    \textbf{Judge model} & \textbf{Aggregate recall} \\
    \midrule
    Sonnet 4.5            & 0.457 \\
    Sonnet 4.6             & 0.539$^{*}$ \\
    Sonnet 5               & 0.497 \\
    Llama 3.3 70B          & 0.525 \\
    GPT-5.6 Sol             & 0.574$^{*}$ \\
    GPT-5.6 Terra           & 0.548$^{*}$ \\
    GPT-5.6 Luna            & 0.567$^{*}$ \\
    GPT-OSS 120B            & 0.476$^{*}$ \\
    \bottomrule
  \end{tabular}
  \caption{Full eight-model Retrieve-Judge leaderboard (composed
  retrieval+judge aggregate recall, 302-pair benchmark). Sonnet 4.5, Sonnet 5,
  and Llama 3.3 70B are measured directly;
  cells marked $^{*}$ are composed from each model's measured per-class judge
  recall and the shared leaf-corrected per-class retrieval recall (Conflict
  0.571, Redundancy 0.846, Similarity 0.480), support-weighted over
  Conflict/Redundancy/Similarity. The retrieval factor is common to every row,
  so it sets the band's absolute level but not the model ordering.}
  \label{tab:leaderboard}
\end{table}

\label{sec:appendix-llama}
The Llama 3.3 70B row also measures prompt transfer: under the Sonnet-tuned
judge prompt its judge-only Conflict recall was 0.654, and a from-scratch
Llama-specific prompt raised it to 0.808 while Conflict precision held near
0.49, a label-boundary limit rather than a wording one.

The confirm-model swap of Section~\ref{sec:eval-results} is in Table~\ref{tab:confirm-swap}.

\begin{table}[ht]
  \centering
  \footnotesize
  \setlength{\tabcolsep}{1.5pt}
  \begin{tabular}{@{}lccccc@{}}
    \toprule
    \textbf{Confirm model} & \textbf{Confl.} & \textbf{Redun.} & \textbf{Simil.} & \textbf{Diff.} & \textbf{Acc.} \\
    \midrule
    Sonnet 4.5 (production) & \textbf{0.885} & 0.842 & 0.682 & 0.789 & \textbf{0.791} \\
    Sonnet 5              & 0.692 & 0.930 & 0.773 & 0.709 & 0.758 \\
    Opus 4.8               & 0.423 & 0.895 & 0.568 & 0.823 & 0.765 \\
    GPT-5.6 Terra           & 0.808 & 0.895 & 0.705 & 0.709 & 0.752 \\
    GPT-5.6 Luna            & 0.808 & 0.877 & 0.705 & 0.623 & 0.699 \\
    GPT-OSS 120B            & 0.731 & 0.912 & 0.727 & 0.543 & 0.656 \\
    Llama 3.3 70B           & 0.846 & 0.982 & 0.659 & 0.474 & 0.629 \\
    \bottomrule
  \end{tabular}
  \caption{Screen-Confirm with its \emph{confirm} model swapped (Haiku screen
  fixed): per-class recall (including Different) and four-class accuracy on the
  302 pairs. Conflict recall is not monotonic in general model strength; the
  Different column tracks each model's ability to walk a false escalation back
  to Different, where Llama is weakest (0.474).}
  \label{tab:confirm-swap}
\end{table}

\section{Negative Ablations}
\label{sec:appendix-negative}

For completeness we list experiments that produced no usable improvement and
are not part of the deployed system; each was run in full before being set aside.

\begin{itemize}
  \item \textbf{Embedding-side retrieval fixes (five, all negative).} Parent-section
  index alongside leaves, LLM-generated context concatenated into the embedding,
  section/document title prepended to embedded text, 1024$\rightarrow$1536
  output dimension, and float32 versus int8 Cohere-side quantization. None
  improved Conflict or Similarity R@10; two regressed it. The retrieval gap is
  structural (chunk boundaries and the leaf filter), not an embedding parameter.
  \item \textbf{Reranking and hypothetical-document queries.} Cohere Rerank
  over the dense top-20 moved per-class R@10 by at most $\pm$0.05 and changed
  no composed cell, so the scored configuration is dense-only. Hypothetical
  partner passages generated by Haiku and embedded as the query (HyDE) landed on
  the right topic but not the annotated partner among many same-topic chunks,
  with no net R@10 gain.
  \item \textbf{Chain-of-thought judge prompt.} A forced reasoning protocol
  before the label lowered Retrieve-Judge Conflict precision on the benchmark
  (0.444$\rightarrow$0.304): it applied the new rule inconsistently across
  near-duplicate phrasings.
  \item \textbf{Long-context, no-retrieval baseline.} Feeding whole documents
  (up to 533K tokens) to the judge scored within noise of quote-only judging on an
  18-pair pilot at roughly 39$\times$ the input tokens per pair.
  \item \textbf{Agentic tool-use judge.} Letting the judge fetch surrounding
  context at its own discretion was a wash on 302 pairs (24 rows fixed, 24 broken)
  at about 3.7$\times$ cost, and introduced a new failure: irrelevant adjacent
  context turning correct Different calls into false Conflicts.
\end{itemize}

\section{Which Fixes Moved Recall}
\label{sec:eval-evolution}
Screen-Confirm's end-to-end row is scored on the production index: for each
benchmark pair the screen reads the anchor document's entire provisions store
(614 provisions on average over the 78 indexed documents; 13{,}352 for the
957-page Departmental Orders) against the section of the compared document
that holds the partner text, and confirm follows. It scores \eeAgg{} aggregate
and \eeConfl{} on Conflict. Table~\ref{tab:e2e-funnel} shows where each gold positive was lost:
7 of 127 positives never reach a model, the screen drops 31, confirm relabels
29, and 20 of 175 Different pairs survive as false positives. Cross-document
pairs recover 0.541; within-document pairs
0.377. Precision with confidence intervals is in
Appendix~\ref{sec:appendix-precision}.

\begin{table}[ht]
  \centering
  \scriptsize
  \setlength{\tabcolsep}{2.5pt}
  \begin{tabular}{@{}lcccccc@{}}
    \toprule
    \textbf{Class} ($n$) & \textbf{Absent} & \textbf{No prov.} & \textbf{Screen} & \textbf{Confirm} & \textbf{Found} & \textbf{Recall} \\
    \midrule
    Confl. (26) & 0 & 0 &  6 &  4 & 16 & 16/26 \\
    Redun. (57) & 3 & 3 &  8 & 11 & 32 & 32/51 \\
    Simil. (44) & 1 & 0 & 17 & 14 & 12 & 12/43 \\
    \midrule
    All (127)   & 4 & 3 & 31 & 29 & 60 & 60/120 \\
    \bottomrule
  \end{tabular}
  \caption{Where each of the 127 gold positives ended on the production data
  path, final run. Absent: its text is not in the index. No prov.\ (no provision): indexed, but distillation
  produced no provision for it. Screen, Confirm: lost at that stage. Found:
  labelled correctly. Recall: over pairs that reached the screen.}
  \label{tab:e2e-funnel}
\end{table}

This stage-by-stage attribution was re-run after each fix, and each run said
what to fix next. The first run lost 67 of 127 positives before
any model call, to three ingest faults: split-PDF parts overwrote each other's
chunks, short chunks were dropped, and distillation batches were truncated.
Fixing them took recall from 0.260 to 0.354 and left 4 positives absent. Two
design faults followed. The screen read about 6{,}000 provisions per call;
windows of about 1{,}500 took recall to 0.449 (Conflict 12\,$\rightarrow$\,17 of
26) at about 2.3$\times$ the run's cost. Confirm judged a provision against a
whole section, so a sentence duplicated inside a longer section read as
Similarity; asking it to label against the matching passage took recall to
\eeAgg{} (Redundancy 27\,$\rightarrow$\,32 of 57). Both cost precision, with
Different pairs surviving confirm rising from 12 to 20 of 175.

\section{Benchmark-Scale Cost and Latency}
\label{sec:appendix-cost}
Table~\ref{tab:cost-latency} gives cost and latency for the 302-pair benchmark run, with the pair given; deployment-scale costs are in Section~\ref{sec:eval-cost}.

\begin{table}[ht]
  \centering
  \small
  \begin{tabular}{lccc}
    \toprule
    \textbf{System} & \textbf{Cost} & \textbf{\$/1k} & \textbf{Latency} \\
    \midrule
    Screen-Confirm  & \$1.95 & \$6.47 & ${\sim}$19\,min$^{\S}$ \\
    Retrieve-Judge  & \$2.13$^{\ddagger}$ & \$7.04 & ${\sim}$28\,min$^{\S}$ \\
    Llama (zero-shot) & \$0.01 & \$0.03 & ${\sim}$2\,min \\
    DeBERTa NLI     & \$0.00 & \$0.00 & ${\sim}$2\,min \\
    \bottomrule
  \end{tabular}
  \caption{\textbf{Benchmark-scale} cost and latency, sequential single
  worker, not production cost. Costs are summed from logged token counts at
  list rate. $^{\ddagger}$Retrieve-Judge output tokens estimated from a
  40-pair sample; with prompt caching its cost falls to ${\sim}$\$0.91.
  $^{\S}$Wall-clock from an earlier 376-pair run of the same pipelines; the
  302-pair run did not log it.}
  \label{tab:cost-latency}
\end{table}

\section{Precision and Confidence Intervals}
\label{sec:appendix-precision}
Screen-Confirm's per-class precision and recall with the pair given (the setting of Table~\ref{tab:perclass-recall}), with 95\% Wilson intervals on recall. Redundancy precision is 1.000 and Similarity's 0.612: every pair the cascade called Redundancy was Redundancy, and its mistakes were Redundancy pairs called Similarity. It did not mislabel any pair as Conflict or Different.

\begin{table}[ht]
  \centering
  \small
  \setlength{\tabcolsep}{3pt}
  \begin{tabular}{@{}lccc@{}}
    \toprule
    \textbf{Class} ($n$) & \textbf{Prec.} & \textbf{Recall} & \textbf{95\% CI} \\
    \midrule
    Conflict (26)    & 0.404 & 0.885 & [0.710, 0.960] \\
    Redundancy (57)  & 1.000 & 0.842 & [0.726, 0.915] \\
    Similarity (44)  & 0.612 & 0.682 & [0.534, 0.800] \\
    Different (175)  & 0.932 & 0.789 & [0.722, 0.843] \\
    \bottomrule
  \end{tabular}
  \caption{Screen-Confirm precision/recall per class, with 95\% Wilson
  intervals on recall.}
  \label{tab:perclass-full}
\end{table}

\noindent Because the benchmark is enriched for the rare classes, precision will be
lower in deployment. Conflict is 26 of the 302 pairs, about 9\%. Given the pair, the
cascade flags 57 Conflicts, 23 right and 34 wrong (precision 0.404). If true Conflicts
are ten times rarer in a real corpus, the 23 become about 2 while the 34 false flags
remain, and precision falls to about 0.06. Recall does not move: 0.885 of true
Conflicts are caught whatever their number.
\section{Prompts}
\label{sec:appendix-prompt-text}
The five prompts below are reproduced verbatim. Variable slots filled at call time are shown in
\slot{braces}; section headers within a prompt are set in bold for readability and em dashes in the originals
are shown as hyphens, with no other change to the text. The Retrieve-Judge judge prompt is the longest because
it carries the label definitions, decision order, and five worked examples; the distillation and screen prompts are short and rely on the provisions list for context.

\subsection{Provision Distillation Prompt (Ingest, Fast Model)}
\label{sec:appendix-prompt-distill}
System prompt; each call receives a batch of up to 100 chunks, each prefixed with its header.
\begin{promptbox}{{Provision distillation (system prompt, fast model)}}\raggedright
You are a policy analyst. Extract every distinct requirement, rule, or obligation from the policy text. Each chunk of input is prefixed with a header like [chunk:<id> | <section> | p.<page>]. Return one bullet per requirement. Each bullet must begin with the exact chunk header from its source chunk, then the requirement as a complete, standalone sentence. Example: [chunk:abc | I > I.A | p.3] Officers must maintain their service weapon. No headings, no grouping, no commentary. Return only prefixed bullets, one per line.
\end{promptbox}

\subsection{Retrieve-Judge Judge Prompt}
\label{sec:appendix-prompt-judge}
\begin{promptbox}{{Retrieve-Judge judge (system prompt)}}\raggedright
You are a senior police policy analyst verifying whether two policy chunks have a genuine relationship that officers need to know about.

These chunks were pre-screened as potentially related. Your job is to CONFIRM the relationship type or REJECT as Different. Decide EVERY boundary by normative CONTENT, not by wording overlap. Apply the DECISION ORDER below in exact order; Similarity is NOT a leftover bucket  -  confirm its three positive conditions.

\textbf{DEFINITIONS} (decide by normative CONTENT, never by shared word count):\\
- Conflict: THE SAME OFFICER in THE SAME SITUATION cannot comply with both  -  obeying one puts them in breach of the other. Time frames, thresholds, or procedures directly contradict.\\
- Redundancy: the two impose materially the SAME obligation  -  same actor, same scope, same threshold  -  so deleting one loses NO requirement. Strip formatting, casing, punctuation, and filler words first; if the only remaining differences are non-normative (reworded, an inserted \textquotedbl{}the\textquotedbl{}, a unicode variant, reordering), it is Redundancy even when the wording is not word-for-word.\\
- Similarity: ALL THREE hold  -  (1) SAME SUBJECT (same act, object, event, or decision; an officer handling one would foreseeably meet the other), (2) COMPATIBLE (both can be obeyed together; neither breaches the other), and (3) MATERIAL NORMATIVE DELTA (each imposes normative content the other lacks: a different duty, actor, recipient, threshold, phase, or added condition). Related rules that neither restate nor contradict each other.\\
- Different: no shared act, object, event, or decision  -  one carries nothing that bears on the other.

\textbf{DECISION ORDER} (apply in this exact order):\\
1. Redundancy check FIRST, by content: same obligation on the same actor/scope/threshold, differing only in non-normative ways? -> Redundancy.\\
2. Conflict check: could obeying one ever breach the other for the same subject in one situation? -> Conflict. A difference in wording, or different-but-compatible details, is NOT a Conflict.\\
3. Different check: does one carry nothing that bears on the other? -> Different.\\
4. Otherwise -> Similarity (same subject + compatible + material normative delta).

\textbf{EXCEPTION HANDLING} (a general rule paired with a narrower passage):\\
- The narrower passage EXPLICITLY subordinates itself and fully dissolves the tension (\textquotedbl{}unless\textquotedbl{}, \textquotedbl{}except\textquotedbl{}, or it names the sub-population it exempts)  -  the tension is resolved -> Different.\\
- It is a conditional override the reader must CROSS-REFERENCE to apply correctly (e.g. a default threshold plus a different threshold when a condition holds) -> Similarity.

\textbf{NOTIFICATION ROUTING}  -  apply this before classifying any communication/reporting pair as Conflict:\\
Requiring notification to Person A (one policy) and Person B (another policy) is NOT a Conflict  -  the officer can notify both. It is Similarity.\\
A Conflict exists only when both policies specify the SAME notification slot but assign it to mutually exclusive recipients.

\textbf{REDUNDANCY vs SIMILARITY}  -  the content test (never decide this by shared word count):\\
Does either passage impose normative content the other lacks  -  a different duty, actor, recipient, threshold, phase, or added condition?\\
- No material delta (only wording/formatting differs) -> Redundancy, even if the two are not word-for-word identical.\\
- At least one material delta -> Similarity, even when long spans of wording are shared.

\textbf{HARD REJECTION CRITERIA} for Conflict (output Similarity or Different if ANY apply):\\
- Different actors (officer vs supervisor vs dispatcher vs records staff) are a MATERIAL DELTA: an otherwise-duplicate rule that assigns the duty to a different actor is Similarity, not Redundancy.\\
  CARVE-OUT  -  Conflict: same procedure, contradictory parameters for situations involving the same officer -> Conflict regardless of unit labels.\\
- Different phases (initial response vs follow-up report vs administrative review)\\
- Same broad topic but different specific situations\\
- General guidance vs specific procedure\\
- One policy sets a standard, the other implements it in a different context\\
- SAME-DOCUMENT PAIRS  -  apply extra caution: apparent contradictions are often complementary steps or sequential procedures. Default to Similarity over Conflict unless a specific numeric value, deadline, or threshold directly contradicts with no possible sequential or role-based reading.

\textbf{AUDIENCE AND UNIT CHECK}  -  before any Conflict:\\
Confirm both chunks apply to the same unit, role, and operational context. If different audiences, units, or operational phases are plausible, classify as Different or Similarity rather than Conflict.

\textbf{FINAL CHECK:}\\
- Conflict: Name the specific scenario where the same officer cannot comply with both simultaneously. If you cannot -> Similarity or Different.\\
- Redundancy: Confirm NO material normative delta remains after stripping formatting and wording. If a delta remains (different actor, threshold, phase, added condition) -> Similarity.

\textbf{EXAMPLES:}

\textbf{Example 1}  -  Conflict (time limit):\\
Chunk A: \textquotedbl{}Officers involved in a use-of-force incident shall submit a written report to their immediate supervisor within four (4) hours of the incident.\textquotedbl{}\\
Chunk B: \textquotedbl{}A Use of Force Report shall be completed and submitted by the involved officer no later than twenty-four (24) hours following any use of force occurrence.\textquotedbl{}\\
Output: \{\textquotedbl{}summary\textquotedbl{}: \textquotedbl{}Conflict\textquotedbl{}, \textquotedbl{}findings\textquotedbl{}: [\{\textquotedbl{}type\textquotedbl{}: \textquotedbl{}Conflict\textquotedbl{}, \textquotedbl{}description\textquotedbl{}: \textquotedbl{}Use of force report deadline is 4 hours (Chunk A) versus 24 hours (Chunk B)  -  the same officer filing the same report faces contradictory deadlines.\textquotedbl{}\}]\}

\textbf{Example 2}  -  Conflict (authorisation threshold):\\
Chunk A: \textquotedbl{}Purchases of evidence-related supplies exceeding \$500.00 require prior written approval from the unit commander.\textquotedbl{}\\
Chunk B: \textquotedbl{}Unit expenditures on evidence supplies up to \$1,000.00 may be authorised by the division supervisor without additional approval.\textquotedbl{}\\
Output: \{\textquotedbl{}summary\textquotedbl{}: \textquotedbl{}Conflict\textquotedbl{}, \textquotedbl{}findings\textquotedbl{}: [\{\textquotedbl{}type\textquotedbl{}: \textquotedbl{}Conflict\textquotedbl{}, \textquotedbl{}description\textquotedbl{}: \textquotedbl{}A \$750 evidence supply purchase requires prior commander approval (Chunk A) but falls within the supervisor's self-authorisation limit (Chunk B)  -  the same officer cannot determine which approval pathway applies.\textquotedbl{}\}]\}

\textbf{Example 3}  -  Redundancy (near-verbatim):\\
Chunk A: \textquotedbl{}Officers must complete a Use of Force Report (Form UF-1) within twenty-four (24) hours of any use of force incident.\textquotedbl{}\\
Chunk B: \textquotedbl{}A Use of Force Report (Form UF-1) shall be completed by the involved officer within twenty-four (24) hours following any use of force occurrence.\textquotedbl{}\\
Output: \{\textquotedbl{}summary\textquotedbl{}: \textquotedbl{}Redundancy\textquotedbl{}, \textquotedbl{}findings\textquotedbl{}: [\{\textquotedbl{}type\textquotedbl{}: \textquotedbl{}Redundancy\textquotedbl{}, \textquotedbl{}description\textquotedbl{}: \textquotedbl{}Both sections require Form UF-1 within 24 hours of a use of force incident using near-identical wording  -  this requirement is duplicated across two documents.\textquotedbl{}\}]\}

\textbf{Example 4}  -  Similarity (notification to different people  -  NOT a Conflict):\\
Chunk A: \textquotedbl{}Officers shall notify their immediate supervisor within one (1) hour of any use of force incident.\textquotedbl{}\\
Chunk B: \textquotedbl{}Following any use of force, the officer shall notify the on-duty watch commander as soon as practicable.\textquotedbl{}\\
Output: \{\textquotedbl{}summary\textquotedbl{}: \textquotedbl{}Similarity\textquotedbl{}, \textquotedbl{}findings\textquotedbl{}: [\{\textquotedbl{}type\textquotedbl{}: \textquotedbl{}Similarity\textquotedbl{}, \textquotedbl{}description\textquotedbl{}: \textquotedbl{}Both policies require post-use-of-force notifications but to different recipients (supervisor vs watch commander)  -  an officer must satisfy both, so awareness of each is operationally relevant.\textquotedbl{}\}]\}

\textbf{Example 5}  -  Different:\\
Chunk A: \textquotedbl{}The Records Unit shall maintain all arrest reports for a minimum of seven (7) years.\textquotedbl{}\\
Chunk B: \textquotedbl{}Officers responding to domestic disturbance calls shall request a supervisor to respond to the scene.\textquotedbl{}\\
Output: \{\textquotedbl{}summary\textquotedbl{}: \textquotedbl{}Different\textquotedbl{}, \textquotedbl{}findings\textquotedbl{}: []\}

\textbf{OUTPUT ONLY valid JSON}  -  no other text:\\
\{\textquotedbl{}summary\textquotedbl{}: \textquotedbl{}Conflict|\allowbreak Redundancy|\allowbreak Similarity|\allowbreak Different\textquotedbl{}, \textquotedbl{}findings\textquotedbl{}: [\{\textquotedbl{}type\textquotedbl{}: \textquotedbl{}Conflict|\allowbreak Redundancy|\allowbreak Similarity\textquotedbl{}, \textquotedbl{}description\textquotedbl{}: \textquotedbl{}concise description of the issue\textquotedbl{}\}]\}

For Different: \{\textquotedbl{}summary\textquotedbl{}: \textquotedbl{}Different\textquotedbl{}, \textquotedbl{}findings\textquotedbl{}: []\}

\textbf{WRITING GUIDANCE}  -  officers review thousands of these:\\
- State the specific contradiction, duplication, or connection in one sentence\\
- Embed section references in parentheses  -  do not start with \textquotedbl{}Section X.X requires...\textquotedbl{}\\
- Do not use: \textquotedbl{}appears in both\textquotedbl{}, \textquotedbl{}creating divergence risk\textquotedbl{}, \textquotedbl{}mutually exclusive\textquotedbl{}, \textquotedbl{}cannot simultaneously comply\textquotedbl{}\\
- Do not add a second sentence unless critical information will not fit in one
\end{promptbox}

\subsection{Screen-Confirm Screen Prompt (Fast Model)}
\label{sec:appendix-prompt-screen}
Sent after the cached, numbered provisions block and the Policy B section (``Policy B section (\textless{}key\textgreater{}):'' followed by the section text). This is the version scored in Section~\ref{sec:eval}; the deployed version at the time of writing is a shorter earlier one with the same output contract.
\begin{promptbox}{{Screen-Confirm screen (fast model)}}\raggedright
Below is a list of numbered provisions from Policy A, then a section from Policy B.\\
This is a fast first-pass screen. Flag every provision that plausibly relates to the Policy B section so it can be re-checked later; when in doubt, INCLUDE it.\\
\textbf{Use these labels:}\\
- Conflict: the provision and the section give directives that cannot both be followed for the same subject  -  one requires what the other forbids, or they set different thresholds/deadlines/numbers for the same thing.\\
- Redundancy: the provision and the section impose the same obligation on the same actor and scope  -  the same requirement restated, differing only in wording or formatting (decide by content, not by shared word count).\\
- Similarity: same subject and compatible, but each carries some normative content the other lacks  -  a different duty, actor, recipient, threshold, phase, or added condition  -  so neither restates nor contradicts the other.\\
\textbf{Examples of what to flag:}\\
- Provision 'Reports are due by the 5th of the month' vs section 'Reports shall be submitted by the 5th of each month' -\textgreater{} Redundancy (same requirement, near-identical text).\\
- Provision 'Officers must always wear a uniform' vs section 'Plain-clothes officers may work in civilian dress' -\textgreater{} Conflict (cannot both hold for the same officers).\\
Skip only provisions that are completely unrelated  -  a different subject with no overlap  -  and output nothing for them.\\
Return a JSON array. Each item: \{\textquotedbl{}provision\_index\textquotedbl{}: int, \textquotedbl{}label\textquotedbl{}: \textquotedbl{}Conflict|\allowbreak Redundancy|\allowbreak Similarity\textquotedbl{}, \textquotedbl{}why\textquotedbl{}: str\}.\\
If nothing matches, return []. Return JSON array only, no other text.
\end{promptbox}

\subsection{Screen-Confirm Confirm Prompt (Strong Model)}
\label{sec:appendix-prompt-confirm}
Cross-document version, scored in Section~\ref{sec:eval}; as above, the deployed version is a shorter earlier one. The screen's label enters only at step~3 as a tie-breaker.
\begin{promptbox}{{Screen-Confirm confirm (strong model), cross-document}}\raggedright
\textbf{Policy A text:}\\
\slot{policy\_a\_text}

\textbf{Policy B section text:}\\
\slot{policy\_b\_text}

Classify the relationship between these two passages. Work in order; do NOT consider any prior system's guess until step 3.\\
\textbf{Step 1}  -  In one clause each, state the specific requirement each passage imposes (who must do what, when, to what threshold). If a passage imposes no requirement, say so.\\
\textbf{Step 2}  -  Using ONLY your step-1 reading, choose the label by CONTENT (never by shared word count), checking in this order:\\
- Redundancy (check FIRST): the two impose materially the SAME obligation  -  same actor, scope, and threshold  -  so deleting one loses no requirement. Strip formatting and wording; if the only differences are non-normative (rewording, casing, an inserted 'the', a unicode variant), it is Redundancy even if not word-for-word.\\
- Conflict: the two requirements CANNOT both be satisfied for the same subject in the same situation. Apply this concrete test  -  would obeying Policy A put someone in breach of Policy B in that same situation? Only 'yes' is a Conflict. Shared topic, overlapping wording, or different-but-compatible details are NOT a Conflict.\\
- Similarity: ALL THREE hold  -  same subject, compatible (both can be obeyed together), and a MATERIAL NORMATIVE DELTA (each imposes content the other lacks: a different duty, actor, recipient, threshold, phase, or added condition). A different actor or threshold is a material delta  -  Similarity, not Redundancy, even when wording is largely shared.\\
- Different: no shared act, object, event, or decision. Finding ONE distinguishing detail does not by itself make a pair Different when both passages still address the same underlying subject  -  that is Similarity, not Different.\\
\textbf{Contrastive examples} (same format, opposite outcomes):\\
- A 'Officers shall document any use of force within 24 hours' vs B 'Supervisors shall review all use-of-force reports' -\textgreater{} Similarity: same subject (use of force), each imposes a different but compatible duty; neither contradicts or restates the other.\\
- A 'Officers shall document any use of force within 24 hours' vs B 'Fleet vehicles shall be refueled at the end of each shift' -\textgreater{} Different: unrelated subjects, no shared requirement.\\
\textbf{Step 3}  -  A first-pass screen guessed: \slot{label} (reason: \slot{why}). Use this only as a tie-breaker: keep your step-2 label as-is; adopt the guess ONLY if you were genuinely undecided between two labels and it breaks the tie. Do not switch to it merely because it was provided.\\
Reply with JSON only:\\
\{\textquotedbl{}confirmed\textquotedbl{}: true/false, \textquotedbl{}label\textquotedbl{}: \textquotedbl{}Conflict|\allowbreak Redundancy|\allowbreak Similarity|\allowbreak Different\textquotedbl{}, \textquotedbl{}why\textquotedbl{}: str, \textquotedbl{}quote\_a\textquotedbl{}: str, \textquotedbl{}quote\_b\textquotedbl{}: str\}\\
confirmed: true if your final label equals the first-pass guess, false otherwise.\\
quote\_a: short verbatim phrase from Policy A most relevant to the finding (1-2 sentences, empty string if Different).\\
quote\_b: short verbatim phrase from Policy B most relevant to the finding (1-2 sentences, empty string if Different).
\end{promptbox}
End-to-end runs (Section~\ref{sec:eval-e2e}) replace the opening instruction
with: \emph{The Policy B section may contain several requirements. First find
the single passage (1--3 sentences) in it that addresses the same subject as
the Policy A text. Classify the relationship between the Policy A text and
THAT passage only; ignore requirements elsewhere in the section.}

\subsection{Screen-Confirm Confirm Prompt, Same-Document Variant}
\label{sec:appendix-prompt-confirm-self}
Used when both passages come from one document (the exception-handling rules differ).
\begin{promptbox}{{Screen-Confirm confirm (strong model), same-document}}\raggedright
Both passages below are drawn from the SAME policy document. Treat them as parts of one governing authority that is meant to be internally consistent.

\textbf{Passage A:}\\
\slot{policy\_a\_text}

\textbf{Passage B:}\\
\slot{policy\_b\_text}

Classify the relationship. Work in order; do NOT consider any prior system's guess until step 3.\\
\textbf{Step 1}  -  In one clause each, state the specific requirement each passage imposes (who must do what, when, under what circumstances).\\
\textbf{Step 2}  -  Using ONLY your step-1 reading, choose the label:\\
- Conflict: the two passages issue directives that CANNOT both be complied with for the same subject under the same circumstances. Apply this concrete test  -  would obeying Passage A put someone in breach of Passage B in that same situation? Only 'yes' is a Conflict.\\
- NOT a conflict (label Different): a general rule paired with a narrower exception that is EXPLICITLY subordinated to it  -  the narrower passage signals it is a carve-out (e.g. 'unless', 'except', 'shall not ... unless', or it names the specific sub-population it exempts from the general rule). The exception refines the rule and both can be followed together. But if the two passages are each free-standing rules that independently reach OPPOSITE determinations about the same subject (neither is written as an exception to the other), that remains a Conflict.\\
- NOT a conflict (label Different): passages that govern different subjects, roles, units, materials, or circumstances, even when they share wording or cite different numbers/thresholds.\\
- Redundancy: the two impose materially the SAME obligation  -  same actor, scope, and threshold  -  differing only in non-normative ways (wording, casing, formatting). Decide by content, not by shared word count.\\
- Similarity: ALL THREE hold  -  same subject, compatible, and a MATERIAL NORMATIVE DELTA (each imposes content the other lacks: a different duty, actor, threshold, phase, or added condition). This INCLUDES a general default plus a conditional override the reader must CROSS-REFERENCE to apply correctly (e.g. a different threshold when some condition holds, neither written as an explicit exception to the other). A different actor or threshold is a material delta  -  Similarity, not Redundancy. Finding ONE distinguishing detail does not by itself make a pair Different when both still address the same subject.\\
- Different: no shared act, object, event, or decision, or an apparent contradiction resolved by an explicit exception or by differing scope as above.\\
\textbf{Contrastive examples} (same format, opposite outcomes):\\
- A 'All prisoners will be handcuffed during transport' vs B 'Juveniles 12 and under shall not be handcuffed unless they pose a threat' -\textgreater{} Different: B is an explicit exception ('unless') carved out of A; both can be followed together.\\
- A 'Use-of-force reports are due within 24 hours' vs B 'Use-of-force reports are due within 72 hours' -\textgreater{} Conflict: two free-standing rules set incompatible deadlines for the same report, neither written as an exception to the other.\\
\textbf{Step 3}  -  A first-pass screen guessed: \slot{label} (reason: \slot{why}). Use this only as a tie-breaker: keep your step-2 label as-is; adopt the guess ONLY if you were genuinely undecided between two labels and it breaks the tie. Do not switch to it merely because it was provided.\\
Reply with JSON only:\\
\{\textquotedbl{}confirmed\textquotedbl{}: true/false, \textquotedbl{}label\textquotedbl{}: \textquotedbl{}Conflict|\allowbreak Redundancy|\allowbreak Similarity|\allowbreak Different\textquotedbl{}, \textquotedbl{}why\textquotedbl{}: str, \textquotedbl{}quote\_a\textquotedbl{}: str, \textquotedbl{}quote\_b\textquotedbl{}: str\}\\
confirmed: true if your final label equals the first-pass guess, false otherwise.\\
quote\_a: short verbatim phrase from Passage A most relevant to the finding (1-2 sentences, empty string if Different).\\
quote\_b: short verbatim phrase from Passage B most relevant to the finding (1-2 sentences, empty string if Different).
\end{promptbox}

\subsection{Baseline Setups}
\label{sec:appendix-prompt-baselines}
\textbf{Llama zero-shot.} \texttt{us.meta.llama3-3-70b-instruct-v1:0} via Bedrock, one call per pair, quote-only, temperature 0, no examples. System instruction:
\begin{promptbox}{{Llama zero-shot baseline (system instruction)}}\raggedright
You classify the relationship between two police policy statements into exactly one of four categories:\\
- Conflict: an officer cannot comply with both at once; they contradict.\\
- Redundancy: they impose the same requirement; one duplicates the other.\\
- Similarity: related and compatible, but each adds something the other lacks.\\
- Different: unrelated; neither bears on the other.\\
Answer with only one word: Conflict, Redundancy, Similarity, or Different.
\end{promptbox}
\textbf{DeBERTa NLI.} \texttt{MoritzLaurer/\allowbreak deberta-v3-large-\allowbreak zeroshot-v2.0}, not fine-tuned, run as zero-shot classification over the concatenated pair with the hypothesis template ``These two policy statements are \{\}.'' and one candidate per label: \emph{in conflict}, \emph{redundant}, \emph{similar}, \emph{unrelated}. The highest entailment score wins; ties break in that order.

\end{document}